# Introducing Variable Importance Tradeoffs into CP-Nets

**Ronen I. Brafman**   **Carmel Domshlak**
Computer Science Dept., Ben-Gurion University
Beer-Sheva 84105, Israel
`brafman,dcarmel@cs.bgu.ac.il`

## Abstract

The ability to make decisions and to assess potential courses of action is a corner-stone of many AI applications, and usually this requires explicit information about the decision-maker's preferences. In many applications, preference elicitation is a serious bottleneck. The user either does not have the time, the knowledge, or the expert support required to specify complex multi-attribute utility functions. In such cases, a method that is based on intuitive, yet expressive, preference statements is required. In this paper we suggest the use of TCP-nets, an enhancement of CP-nets, as a tool for representing, and reasoning about qualitative preference statements. We present and motivate this framework, define its semantics, and show how it can be used to perform constrained optimization.

## 1 INTRODUCTION

The ability to make decisions and to assess potential courses of action is a corner-stone of many AI applications, including expert systems, autonomous agents, decision-support systems, recommender systems, configuration software, and constrained optimization applications. To make good decisions, we must be able to assess and compare different alternatives. Sometimes, this comparison is performed implicitly, as in many recommender systems. However, in many cases explicit information about the decision-maker's preferences is required.

Utility functions are an ideal tool for representing and reasoning with preferences. However, they can be very difficult to elicit, and the effort required is not always possible or justified. Instead, one should resort to other, more qualitative forms of preference representation. Ideally, this qualitative information should be easily obtainable from the user by non-intrusive means. That is, we should be able to generate it from natural and relatively simple statements about preferences obtained from the user, and this elicitation process should be amenable to automation. In addition, automated reasoning with this representation should be feasible and efficient.

One relatively recent framework for preference representation that addresses these concerns is that of *Conditional Preference Networks* (CP-nets) [1, 2]. In CP-nets, the decision maker is asked to describe how her preference over the values of one variable depends on the value of other variables. For example, she may state that her preference for a dessert depends on the value of the main-course as well as whether or not she had an alcoholic beverage. Her choice of an alcoholic beverage depends on the main course and the time of day. This information is described by a graphical structure in which the nodes represent variables of interest and the edges represent dependence relations between the variables. Each node is annotated with a *conditional preference table* (CPT) describing the user's preference over alternative values of this node given different values of the parent nodes. CP-nets capture a class of intuitive and useful natural language statements of the form "I prefer the value $x_0$ for variable $X$ given that $Y = y_0$ and $Z = z_0$". Such statements do not require complex introspection nor a quantitative assessment.

In [1] it was observed that there is another class of preferential statements, not captured by the CP-net model, that is no less intuitive or important. These statements have the following form: "It is more important to me that the value of $X$ be high than that the value of $Y$ be high." We call these *relative importance* statements. For instance, one might say "The length of the journey is more important to me than the choice of airline". A more refined notion of importance, though still intuitive and easy to communicate, is that of *conditional relative importance*: "The length of the journey is more important to me than the choice of airline provided that I am lecturing the following day. Otherwise, the choice of airline is more important." This latter statement is of the form: "A better assignment for $X$ is more important than a better assignment for $Y$ given that $Z = z_0$."

Notice that information about relative importance is different from information about independence. In the example above, my preference for an airline does not depend on the duration of the journey because, e.g., I compare airlines based on their service, security levels and the quality of their frequent flyer program.

In this paper we show that enriching a CP-net based preferential relation by adding such statements may have a significant impact on both the consistency of the specified relation, and the reasoning about it. Likewise, we show that the internal structure of such a "mixed" preferential statement set can be exploited in order to achive efficiency in both consistency testing and in preferential reasoning. In particular, we present an extension of CP-nets, which we call TCP-nets (for *tradeoffs-enhanced* CP-nets), and show how they can be used to compute optimal outcomes given constraints. TCP-nets capture both information about conditional independence and about conditional relative importance. Thus, they provide a richer framework for representing user preferences, allowing stronger conclusions to be drawn, yet remain committed to the use of intuitive, qualitative information as their source.

This paper is organized as follows. In Section 2 we describe the notions underlying TCP-nets: preference relations, preferential independence, and relative importance. In Section 3 we define TCP-nets, and provide a number of examples. In Section 4 we define the semantics of TCP-nets and discuss the conditions for the consistency of the specified preferential orders. In Section 5 we show how TCP-nets can be used to perform constrained optimization. We conclude with a discussion of future work in Section 6. Proofs and a discussion of the TCP-nets applicability to the configuration problems appear in [3].

## 2 PREFERENCE ORDERS, INDEPENDENCE, AND RELATIVE IMPORTANCE

In this section we describe the ideas underlying TCP-nets: preference orders, preferential independence and conditional preferential independence, as well as relative importance and conditional relative importance.

### 2.1 PREFERENCE AND INDEPENDENCE

A *preference relation* is a total pre-order (a *ranking*) over a set of outcomes. Given two outcomes $o, o'$, we write $o \succeq o'$ to denote that $o$ is at least as preferred as $o'$ and we write $o \succ o'$ to denote that $o$ is strictly more preferred than $o'$. The types of outcomes we are concerned with consist of possible assignments to some set of variables. More formally, we assume some given set $V = \{X_1, \ldots, X_n\}$ of variables with corresponding domains $\mathcal{D}(X_1), \ldots, \mathcal{D}(X_n)$. The set of possible outcomes is then $\mathcal{D}(X_1) \times \cdots \times \mathcal{D}(X_n)$. For example, in the context of the problem of configuring a personal computer (PC), the variables may be *processor type, screen size, operating system* etc., where *screen size* has the domain {*17in, 19in, 21in*}, *operating system* has the domain {*LINUX, Windows98, WindowsXP*}, etc. Each assignment to the set of variables specifies an outcome — a particular PC configuration. Thus, a preference ordering over these outcomes specifies a ranking over possible PC configurations.

The number of possible outcomes is exponential in $n$, while the set of possible total orders on them is doubly exponential in $n$. Therefore, explicit specification and representation of a ranking is not realistic. We must implicitly describe this preference relation. Often, the notion of preferential independence plays a key role in such representations. Intuitively, $\mathbf{X}$ and $\mathbf{Y} = \mathbf{V} - \mathbf{X}$ are *preferentially independent* if for all assignments to $\mathbf{Y}$, our preference over $\mathbf{X}$ values are identical. More formally, let $\mathbf{x}_1, \mathbf{x}_2 \in \mathcal{D}(\mathbf{X})$ for some $\mathbf{X} \subseteq \mathbf{V}$ (where we use $\mathcal{D}(\cdot)$ to denote the domain of a set of variables as well), and let $\mathbf{y}_1, \mathbf{y}_2 \in \mathcal{D}(\mathbf{Y})$, where $\mathbf{Y} = \mathbf{V} - \mathbf{X}$. We say that $\mathbf{X}$ is *preferentially independent* of $\mathbf{Y}$ iff, for all $\mathbf{x}_1, \mathbf{x}_2, \mathbf{y}_1, \mathbf{y}_2$ we have that

$$\mathbf{x}_1\mathbf{y}_1 \succeq \mathbf{x}_2\mathbf{y}_1 \text{ iff } \mathbf{x}_1\mathbf{y}_2 \succeq \mathbf{x}_2\mathbf{y}_2$$

For example, in our PC configuration example, the user may assess *screen size* to be preferentially independent of *processor type* and *operating system*. This could be the case if the user always prefers a larger screen to a smaller screen, no matter what the processor or the OS are.

Preferential independence is a strong property, and therefore, less common. A more refined notion is that of conditional preferential independence. Intuitively, $\mathbf{X}$ and $\mathbf{Y}$ are *conditionally preferentially independent* given $\mathbf{Z}$ if for every fixed assignment to $\mathbf{Z}$, the ranking of $\mathbf{X}$ values is independent of the value of $\mathbf{Y}$. Formally, let $\mathbf{X}, \mathbf{Y}$ and $\mathbf{Z}$ be a partition of $\mathbf{V}$ and let $\mathbf{z} \in \mathcal{D}(\mathbf{Z})$. $\mathbf{X}$ and $\mathbf{Y}$ are *conditionally preferentially independent* given $\mathbf{z}$ iff, for all $\mathbf{x}_1, \mathbf{x}_2, \mathbf{y}_1, \mathbf{y}_2$ we have that

$$\mathbf{x}_1\mathbf{y}_1\mathbf{z} \succeq \mathbf{x}_2\mathbf{y}_1\mathbf{z} \text{ iff } \mathbf{x}_1\mathbf{y}_2\mathbf{z} \succeq \mathbf{x}_2\mathbf{y}_2\mathbf{z}$$

$\mathbf{X}$ and $\mathbf{Y}$ are conditionally preferentially independent given $\mathbf{Z}$ if they are conditionally preferentially independent given any assignment $\mathbf{z} \in \mathcal{D}(\mathbf{Z})$. Returning to our PC example, the user may assess *operating system* to be independent of all other features given *processor type*. That is, it always prefers LINUX given an AMD processor and Windows98 given an Intel processor (e.g., because he might believe that Windows98 is optimized for the Intel processor, whereas LINUX is otherwise better). Note that the notions of preferential independence and conditional preferential independence are among a number of standard notions of independence in multi-attribute utility theory [5].

## 2.2 RELATIVE IMPORTANCE

Although statements of preferential independence are natural and useful, the orderings obtained by relying on them alone are relatively weak. To understand this, consider two preferentially independent boolean attributes $A$ and $B$ with values $a_1, a_2$ and $b_1, b_2$, respectively. If $A$ and $B$ are preferentially independent, then we can specify a preference order over $A$ values, say $a_1 \succ a_2$, independently of the value of $B$. Similarly, our preference over $B$ values, say $b_1 \succ b_2$, is independent of the value of $A$. From this we can deduce that $a_1 b_1$ is the most preferred outcome and $a_2 b_2$ is the least preferred outcome. However, we do not know the relative order of $a_1 b_2$ and $a_2 b_1$. This is typically the case when we consider independent variables: We can rank each one given a fixed value of the other, but often, we cannot compare outcomes in which both values are different. One type of information that can address some (though not necessarily all) such comparisons is information about relative importance. For instance, if we say that $A$ is more important than $B$ then this means that we prefer to reduce the value of $B$ rather than reduce the value of $A$. In that case, we know that $a_1 b_2 \succ a_2 b_1$, and we can (totally) order the set of outcomes as $a_1 b_1 \succ a_1 b_2 \succ a_2 b_1 \succ a_2 b_2$.

Returning to our PC configuration example, suppose that *operating system* and *processor type* are independent attributes. We might say that *processor type* is more important than *operating system*, e.g, because we believe that the effect of the processor's type on system performance is more significant than the effect of the operating system.

Formally, let a pair of variables $X$ and $Y$ be preferentially independent given $\mathbf{W} = \mathbf{V} - \{X, Y\}$. We say that $X$ is *more important* than $Y$, denoted by $X \triangleright Y$, if for every assignment $\mathbf{w} \in \mathcal{D}(\mathbf{W})$ and for every $x_i, x_j \in \mathcal{D}(X)$, $y_a, y_b \in \mathcal{D}(Y)$, such that $x_i \succ x_j$ given $\mathbf{w}$ and $y_b \succ y_a$ given $\mathbf{w}$, we have that:

$$x_i y_a \mathbf{w} \succ x_j y_b \mathbf{w}.$$

For instance, when both $X$ and $Y$ are binary variables, and $x_1 \succ x_2$ and $y_1 \succ y_2$ hold given $\mathbf{w}$, then $X \triangleright Y$ iff we have $x_1 y_2 \mathbf{w} \succ x_2 y_1 \mathbf{w}$ for all $\mathbf{w} \in \mathcal{D}(\mathbf{W})$. Notice that this is a strict notion of importance – any reduction in $Y$ is preferred to any reduction in $X$. Clearly, this idea can be refined by providing an actual ordering over elements of $\mathcal{D}(XY)$. We have decided not to pursue this option farther because it is less natural to specify. However, our results generalize to such specifications as well. In addition, one can consider relative importance assessments among more than two variables. However, we feel that such statements are somewhat artificial and less natural to articulate.

Relative importance information is a natural enhancement of independence information. It retains the property we value so much: it corresponds to statements that a naive user would find simple and clear to evaluate and articulate. Moreover, it can be generalized naturally to a notion of *conditional relative importance*. For instance, suppose that the relative importance of *processor type* and *operating system* depends on the primary usage of the PC. For example, when the PC is used primarily for graphical applications, then the choice of an operating system is more important than that of a processor because certain important software packages for graphic design are not available on LINUX. However, for other applications, the processor type is more important because applications for both Windows and LINUX exist. Thus, we say that $X$ is more important than $Y$ given $\mathbf{z}$ if we always prefer to reduce the value of $Y$ rather than the value of $X$ when $\mathbf{z}$ holds.

Formally, let $X, Y, \mathbf{W}$ be as above, and let $\mathbf{Z} \subseteq \mathbf{W}$. We say that $X$ is *more important* than $Y$ given an assignment $\mathbf{z} \in \mathcal{D}(\mathbf{Z})$ (*ceteris paribus*) iff, for any assignment $\mathbf{w}$ on $\mathbf{W} = \mathbf{V} - (\{X, Y\} \cup \mathbf{Z})$ we have:

$$x_i y_a \mathbf{z} \mathbf{w} \succ x_j y_b \mathbf{z} \mathbf{w}$$

whenever $x_i \succ x_j$ given $\mathbf{z}\mathbf{w}$ and $y_b \succ y_a$ given $\mathbf{z}\mathbf{w}$. We denote this relation by $X \triangleright_{\mathbf{z}} Y$. Finally, if for some $\mathbf{z} \in \mathcal{D}(\mathbf{Z})$ we have that either $X \triangleright_{\mathbf{z}} Y$, or $Y \triangleright_{\mathbf{z}} X$, then we say that the relative importance of $X$ and $Y$ is conditioned on $\mathbf{Z}$, and write $\mathcal{RI}(X, Y, \mathbf{Z})$.

## 3 TCP NETS

TCP-nets (for *CP-nets with tradeoffs*) is an extension of CP-nets [2] that encodes (conditional) preferential independence and (conditional) relative importance statements. We use this graph-based representation for two reasons: First, it is an intuitive visual representation of preference independence and relative importance statements. Second, the structure of the graph has important consequences to issues such as consistency and complexity of reasoning. For instance, as we show later, when this structure is "acyclic" (for a suitable definition of this notion!), then the preference statements contained in the graph are consistent – that is, there is a total pre-order that satisfies them.

TCP-nets are annotated graphs with three types of edges. The nodes of a TCP-net correspond to the problem variables $\mathbf{V}$. The first type of (directed) edge captures preferential dependence, i.e., an edge from $X$ to $Y$ implies that the user has different preferences over $X$ values given different values of $Y$. The second (directed) edge type captures relative importance relations. The existence of such an edge from $X$ to $Y$ implies that $X$ is more important than $Y$. The third (undirected) edge type captures conditional importance relations: Such an edge between nodes $X$ and $Y$ exists if there exists some $\mathbf{Z}$ for which $\mathcal{RI}(X, Y, \mathbf{Z})$ holds.

Like in CP-nets, each node $X$ in a TCP-net is annotated with a *conditional preference table* (CPT). This table associates a preferences over $\mathcal{D}(X)$ for every possible value

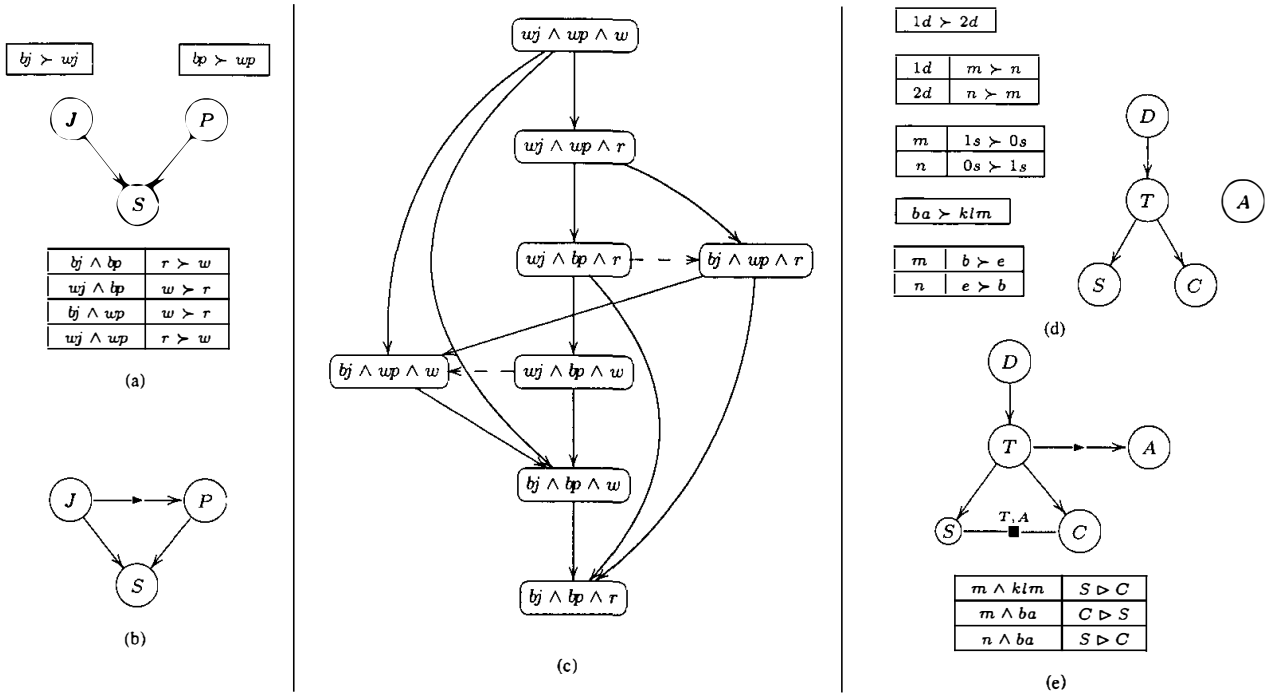

Figure 1: Illustrations: "Evening Dress" CP-net (a) and TCP-net (b); (c) Preferential orderings for (a) and (b); "Flight to the USA" CP-net (d) and TCP-net (e)

assignment to the parents of $X$ (denoted $Pa(X)$). In addition, in TCP-nets, each undirected edge is annotated with a *conditional importance table* (CIT). The CIT associated with such an edge $(X, Y)$ describes the relative importance of $X$ and $Y$ given the value of the conditioning variables.

Formally, a TCP-net $\mathcal{N}$ is a tuple $\langle \mathbf{V}, \mathsf{cp}, \mathsf{i}, \mathsf{ci}, \mathsf{cpt}, \mathsf{cit} \rangle$:

1. $\mathbf{V}$ is a set of nodes, corresponding to the problem variables $\{X_1, \ldots, X_n\}$.

2. cp is a set of directed cp-*arcs* $\{\alpha_1, \ldots, \alpha_k\}$ (where cp stands for *conditional preference*). A cp-arc $\overrightarrow{(X_i, X_j)}$ belongs to $\mathcal{N}$ iff the preferences over the values of $X_j$ depend on the actual value of $X_i$.

3. i is a set of directed i-*arcs* $\{\beta_1, \ldots, \beta_l\}$ (where i stands for *importance*). An i-arc $\overrightarrow{(X_i, X_j)}$ belongs to $\mathcal{N}$ iff $X_i \triangleright X_j$.

4. ci is a set of undirected ci-*arcs* $\{\gamma_1, \ldots, \gamma_m\}$ (where ci stands for *conditional importance*). A ci-arc $(X_i, X_j)$ belongs to $\mathcal{N}$ iff we have $\mathcal{RI}(X_i, X_j, \mathbf{Z})$ for some $\mathbf{Z} \subseteq \mathbf{V} - \{X_i, X_j\}$.

5. cpt associates a CPT with every node $X \in \mathbf{V}$. A CPT is from $\mathcal{D}(Pa(X))$ (i.e., assignment's to $X$'s parent nodes) to total pre-orders over $\mathcal{D}(X)$.

6. cit associates with every ci-arc $(X_i, X_j)$ a subset $\mathbf{Z}$ of $\mathbf{V} - \{X_i, X_j\}$ and a mapping from a subset of $\mathcal{D}(\mathbf{Z})$ to total orders over the set $\{X_i, X_j\}$. We call $\mathbf{Z}$ the *selector set* of $(X_i, X_j)$ and denote it by $\mathcal{S}(X_i, X_j)$.[1]

---

[1] Naturally we expect this set $\mathbf{Z}$ to be the minimal context upon which the relative importance between $X_i$ and $X_j$ depends.

A CP-net [2] is simply a TCP-net in which the sets i and ci (and therefore cit) are empty, and that every node $X \in \mathbf{V}$ is independent of all other nodes given $Pa(X)$. In the rest of this section we provide examples of TCP-net. For simplicity of presentation, all variables in these examples are binary, although the semantics of both CP-net and TCP-net is defined with respect to arbitrary finite domains.

**Example 1** (Evening Dress) Figure 1(a) illustrates another CP-net that expresses my preferences over an evening dress. This network consists of three variables $J$, $P$, and $S$, standing for the jacket, pants, and shirt, respectively. I unconditionally prefer black to white as a color for both the jacket and the shirt, while my preference between the red and white shirts is conditioned by the *combination* of jacket and pants: If they are of the same color, then a white shirt will make my dress too colorless, thus I prefer a red shirt. Otherwise, if the jacket and the pants are of different colors, then a red shirt will probably make my evening dress too flashy, thus I prefer a white shirt. The solid lines in Figure 1(c) presents the corresponding preference relation over the outcomes. The top and the bottom elements are the worst and the best outcomes, respectively. Arrows are directed from less preferred to more preferred outcomes.

In turn, Figure 1(b) displays a TCP-net that extends this CP-net by adding an i-arc from $J$ to $P$, i.e., having black jacket is absolutely more important than having black pants. This induces additional relations among outcomes, captured by the dashed lines in Figure 1(c). The reader may rightfully ask whether this statement of importance is not redundant, since, according to my preference, it seems that I will always wear a completely black suit. However, while my preference is clear, it may definitely be the case that not all the outcomes are feasible at the moment of the actual decision. In particular, it is possible that I will find clean only my velvet black jacket, my velvet white pants, my silk white jacket, and my silk black pants. Since, in my opinion, mixing velvet and silk is simply unacceptable, I will have to compromise, and to wear either the black (velvet) jacket with the white (velvet) pants, or the white (silk) jacket with the black (silk) pants. In this case, my preference for wearing the preferred jacket to wearing the preferred pants determines higher desirability for the velvet combination. Now, if my wife will have to prepare my evening dress while I am late at work writing this thesis, having this information will help her to choose the most preferred, *available* evening dress for me.

**Example 2** (Flight to the USA) Figure 1(d) illustrates a CP-net that capture my preference over the flight options to a conference in USA from Israel. This network consists of five variables:

<u>D</u>ay of the Flight  The variable $D$ distinguishes between flights leaving a day ($D = 1d$) and two days ($D = 2d$) before the conference, respectively. Since I am married, and I am really busy with my work, I prefer to leave on the day before the conference.

<u>A</u>irline  The variable $A$ represents the airline. I prefer to fly with British Airways ($C = ba$) to KLM ($C = klm$).

Departure <u>T</u>ime  The variable $T$ distinguishes between morning/noon ($T = m$) and evening/night ($T = n$) flights. Among flights leaving two days before the conference I prefer an evening/night flight, because it will allow me to work longer at the day of the flight. However, among flights leaving a day before the conference I prefer a morning/noon flight, because I would like to have a few hours before the conference opening in order to take a rest in the hotel.

<u>S</u>top–over  The variable $S$ distinguishes between direct ($S = 0s$) and indirect ($S = 1s$) flights, respectively. I am a smoker, and on day flights I am awake most of the time. Thus, I prefer to have a stop-over in Europe. However, on night flights I sleep, thus I prefer a direct flight since they are shorter.

Seating <u>C</u>lass  The variable $C$ stands for the sitting type. On a night flight, I prefer to sit in the economy class ($C = e$) (I don't care where I sleep, and these seats are significantly cheaper), while on a day flight I prefer to pay for a seat in business class ($C = b$) (I'll be awake so I better have a good seat, good food, good wine).

The CP-net in Figure 1(d) captures all these preferential statements, and the underlying preferential dependencies, while Figure 1(e) presents a TCP-net that extends this CP-net to capture relative importance relations between some parameters of the flight. First, there is an i-arc from $T$ to $A$, since getting more suitable flying time is more important for me than getting the preferred airlines company. Second, there is a ci-arc between $S$ and $C$, where the relative importance of $S$ and $C$ depends on the values of $T$ and $A$:

1. On a KLM day flight, an intermediate stop in Amsterdam is more important to me than sitting in business class (I feel that KLM's business class does not have a good cost/performance ratio, while visiting a casino in Amsterdam's airport sounds to me like a good idea.)

2. For a British Airways nignt flight, the fact that the flight is direct is more important to me than getting a cheaper economy seat (I am ready to pay for a seat in business class, in order not to spend even one minute in Heathrow airport at night).

3. On a British Airways day flight, seating in the business class is more important to me than having a short intermediate break (it is hard to find a nice smoking area in Heathrow).

The CIT of this ci-arc is also presented in Figure 1(e).

## 4  SEMANTICS AND CONSISTENCY

The semantics of a TCP-net is straightforward, and is defined in terms of the set of preference rankings that are consistent with the set of constraints imposed by its preference and importance information. We use $\succ_{\mathbf{u}}^{X}$ to denote the preference relation over the values of $X$ given an assignment $\mathbf{u}$ to $\mathbf{U} \supseteq Pa(X)$.

**Definition 1**  Let $\mathcal{N}$ be a TCP-net over a set of variables $\mathbf{V}$.

1. Let $\mathbf{W} = \mathbf{V} - \{X\} \cup Pa(X)$ and let $\mathbf{p} \in \mathcal{D}(Pa(X))$. A preference ranking $\succ$ satisfies $\succ_{\mathbf{p}}^{X}$ iff $x_i \mathbf{p} \mathbf{w} \succ x_j \mathbf{p} \mathbf{w}$, for each $\mathbf{w} \in \mathcal{D}(\mathbf{W})$, when $x_i \succ_{\mathbf{p}}^{X} x_j$ holds.

2. A preference ranking $\succ$ satisfies the CPT of $X$ iff it satisfies $\succ_{\mathbf{p}}^{X}$ for each assignment $\mathbf{p}$ of $Pa(X)$.

3. A preference ranking $\succ$ satisfies $X \rhd Y$ iff for every $\mathbf{w} \in \mathcal{D}(\mathbf{W})$ s.t. $\mathbf{W} = \mathbf{V} - \{X, Y\}$, $x_i y_a \mathbf{w} \prec x_j y_b \mathbf{w}$ whenever $x_i \succ_{\mathbf{w}}^{X} x_j$ and $y_b \succ_{\mathbf{w}}^{Y} y_a$.

4. A preference ranking $\succ$ satisfies $X \rhd_{\mathbf{z}} Y$ iff for every $\mathbf{w} \in \mathcal{D}(\mathbf{W})$ s.t. $\mathbf{W} = \mathbf{V} - \{X, Y\} \cup \mathbf{Z}$, $x_i y_a \mathbf{z} \mathbf{w} \prec x_j y_b \mathbf{z} \mathbf{w}$ whenever $x_i \succ_{\mathbf{z} \mathbf{w}}^{X} x_j$ and $y_b \succ_{\mathbf{z} \mathbf{w}}^{Y} y_a$.

5. A preference ranking $\succ$ satisfies the CIT of the ci-arc $(X, Y)$ if it satisfies $X \rhd_{\mathbf{z}} Y$ whenever an entry in the table conditioned of $\mathbf{z}$ ranks $X$ as more important.

A preference ranking $\succ$ satisfies a TCP-net $\mathcal{N}$ iff it: (i) satisfies every CPT in $\mathcal{N}$; (ii) satisfies $X \rhd Y$ for every i-arc $(X_i, X_j)$ in $\mathcal{N}$; (iii) satisfied every CIT in $\mathcal{N}$. A TCP-net is *satisfiable* iff there is some ranking $\succ$ that satisfies it. Finally, $o \succ o'$ is *implied* by a TCP-net iff it holds in all preference rankings that satisfy this TCP-net.

**Lemma 1 (Transitivity)**  *If $o \succ o'$ and $o' \succ o''$ are implied by a TCP-net, then so is $o \succ o''$.*

We now define two types of directed graphs that are induced by a TCP-net $\mathcal{N}$.

**Definition 2** $\mathcal{N}$'s *dependency graph* contains all nodes and directed edges of $\mathcal{N}$ (i.e., the cp-arcs and the i-arcs)) as well as the edges $(X_k, X_i)$ and $(X_k, X_j)$ for every ci-arc $(X_i, X_j)$ in $\mathcal{N}$ and every $X_k \in \mathcal{S}(X_i, X_j)$.

Let $\mathcal{S}(\mathcal{N})$ be the union of all selector sets of $\mathcal{N}$. Given an assignment **w** to $\mathcal{S}(\mathcal{N})$, the **w**-*directed* graph of $\mathcal{N}$ contains all nodes and directed edges of $\mathcal{N}$ and the edge from $X_i$ to $X_j$ if $(X_i, X_j)$ is a ci-arc of $\mathcal{N}$ and the CIT for $(X_i, X_j)$ specifies that $X_i \triangleright X_j$ given **w**.

**Definition 3** A TCP-net $\mathcal{N}$ is *conditionally acyclic* if its induced dependency graph is acyclic and for every assignment **w** to $\mathcal{S}(\mathcal{N})$, the induced **w**-directed graphs are acyclic.

**Theorem 1** *Every conditionally acyclic TCP-net is satisfiable.*

Verifying conditional acyclicity requires verifying two properties. The verification of acyclicity of the dependency graph is simple. Naive verification of the acyclicity of every **w**-directed graph can require time exponential in the combined size of the selector sets. Following we show some sufficient and/or neccessary conditions for the **w**-directed graphs acyclicity that are much easier to check.

Let $\mathcal{N}$ be a TCP-net. If $\mathcal{N}$ contains directed cycles, then surely both the induced dependency graph and every **w**-directed graph is cyclic. Since such directed cycles are simple to detect, let us assume that they do not arise in $\mathcal{N}$. Next, note that if there are no cycles in the undirected graph induced by $\mathcal{N}$ (i.e., the graph obtained from $\mathcal{N}$ by removing the direction of its directed edges) then clearly all **w**-directed graphs are acyclic. Again, this case too is quite simple to check. Finally, if there are undirected cycles, but each such cycle, when projected back to $\mathcal{N}$, contains directed arcs in different directions, then all **w**-directed graphs are still acyclic. This latter sufficient condition can be checked in (low) polynomial time.

We are left with the situation that $\mathcal{N}$ contains sets $\mathcal{A}$ of edges that form a cycle in the induced undirected graph, not all of these edges are directed, yet all the directed edges point in the same direction (i.e., clockwise or counterclockwise). We call these *semi-directed* cycles, and focus on their investigation in the rest of this section.

Each assignment **w** to the selector sets of ci-arcs in a semi-directed cycle $\mathcal{A}$ induces a direction to all these arcs. We say that $\mathcal{A}$ is *conditionally acyclic* if under no such assignment **w** do we obtain a directed cycle from $\mathcal{A}$. Otherwise, $\mathcal{A}$ is *conditionally directed*. Our first observation is that if all semi-directed cycles in $\mathcal{N}$ are conditionally acyclic, then so is $\mathcal{N}$. Let $\mathcal{S}(\mathcal{A})$ be the union of the selector sets of all ci-arcs in $\mathcal{A}$. The time required to check for the conditional acyclicity of a semi-directed cycle $\mathcal{A}$ is exponential in the size of $\mathcal{S}(\mathcal{A})$. Thus, if $\mathcal{S}(\mathcal{A})$ is small for each semi-directed cycle $\mathcal{A}$ in the network, then conditionally acyclicity can be checked for quickly. In fact, often we can determine that a semi-directed cycle is conditionally directed/acyclic even more efficiently.

**Lemma 2** *Let $\mathcal{A}$ be a semi-directed cycle in $\mathcal{N}$. If $\mathcal{A}$ is conditionally acylic then it contains a pair of ci-arcs $\gamma_i, \gamma_j$ such that $\mathcal{S}(\gamma_i) \cap \mathcal{S}(\gamma_j) \neq \emptyset$.*

In other words, if the selector sets of the ci-arcs in $\mathcal{A}$ are all pairwise disjoint, then $\mathcal{A}$ is conditionally directed. Thus, Lemma 2 provides a necessary condition for conditional acyclicity of $\mathcal{A}$ that can be checked in time polynomial in the number of variables.

**Lemma 3** *$\mathcal{A}$ is conditionally acyclic if it contains a pair of ci-arcs $\gamma_i, \gamma_j$ such that either:*

*(a) $\mathcal{A}$ contains directed edges and for each assignment **w** to $\mathcal{S}(\gamma_i) \cap \mathcal{S}(\gamma_j)$, $\gamma_i$ or $\gamma_j$ can be converted into an i-arc that violates the direction of the directed edges of $\mathcal{A}$.*

*(b) All edges in $\mathcal{A}$ are undirected and for each assignment **w** to $\mathcal{S}(\gamma_i) \cap \mathcal{S}(\gamma_j)$, $\gamma_i$ and $\gamma_j$ can be converted into i-arcs that point in opposite directions w.r.t. $\mathcal{A}$.*

Lemma 3 provides a sufficient condition for conditional acyclicity of $\mathcal{A}$ that can be checked in time exponential in the maximal size of selector set intersection for a pair of ci-arcs in $\mathcal{A}$. Note that the TCP-net size is at least of this complexity (because of the CITs description), thus checking this condition is only linear in the size of the network.

**Lemma 4** *Let $\mathsf{shared}(\mathcal{A})$ be the union of all pairwise intersections of the selector sets of the ci-arcs in $\mathcal{A}$:*

$$\mathsf{shared}(\mathcal{A}) = \bigcup_{\gamma_i, \gamma_j \in \mathcal{A}} \mathcal{S}(\gamma_i) \cap \mathcal{S}(\gamma_j)$$

*If $\mathcal{A}$ contains some cp or i arcs, then $\mathcal{A}$ is conditionally acyclic if and only if, for each assignment $\pi$ on $\mathsf{shared}(\mathcal{A})$, there exists a ci-arc $\gamma_\pi \in \mathcal{A}$ that, given $\pi$, can be converted into an i-arc that violates the direction of $\mathcal{A}$.*

*Otherwise, if $\mathcal{A}$ consists only of ci-arcs, then $\mathcal{A}$ is conditionally acyclic if and only if, for each assignment $\pi$ on $\mathsf{shared}(\mathcal{A})$, there exist two ci-arcs $\gamma_\pi^1, \gamma_\pi^2 \in \mathcal{A}'$ that, given $\pi$, can be converted into i-arcs that disagree on the direction with respect to $\mathcal{A}$.*

In general, the size of $\mathsf{shared}(\mathcal{A})$ is $O(|\mathbf{V}|)$, thus checking the (necessary and sufficient) condition provided by Lemma 4 is generally hard. However, $|\mathsf{shared}(\mathcal{A})| \leq |\mathcal{S}(\mathcal{A})|$. Therefore, checking this condition is more efficient than checking the naive one. Likewise, restricting the size of $\mathsf{shared}(\mathcal{A})$ (in order to ensure polynomial time consistency verification) will leave us with a much richer set of TCP-nets than restricting the size of $\mathcal{S}(\mathcal{A})$.

## 5 PREFERENTIAL CONSTRAINT-BASED OPTIMIZATION

One of the central properties of the original CP-net model that was presented in [2] is that, given an acyclic CP-net $\mathcal{N}$ and a partial assignment $\pi$ on its variables, it is simple to determine an outcome consistent with $\pi$ that is *preferentially optimal* with respect to $\mathcal{N}$. The corresponding procedure is as follows: Traverse the variables in some topological order induced by $\mathcal{N}$ and set each unassigned variable to its most preferred value given its parents' values. Our immediate observation is that this procedure works *as is* also for conditionally acyclic TCP-nets: The relative importance relations do not play a role in this case, and the network is traversed according to a topological order induced by the CP-net part of the given TCP-net.

This strong property of optimization with respect to the acyclic CP-nets (and the conditionally acyclic TCP-nets) does not hold if some of the TCP-net variables are mutually constrained by a set of hard constraints, $C$. In this case, determining the set of Pareto-optimal[2] feasible outcomes is not trivial. For the acyclic CP-nets, a branch and bound algorithm for determining the optimal feasible outcomes was introduced in [1]. This algorithm has the important *anytime* property — once an outcome is added to the current set of non-dominated outcomes, it is never removed. In this algorithm, variables are instantiated according to a topological ordering. Thus, more important variables, i.e., variables that are "higher-up" in the network, are assigned values first.

Figure 2 presents our extension/modification of that algorithm to conditionally acyclic TCP-nets which retains the *anytime* property. The key difference between processing acyclic CP-net and conditionally acyclic TCP-net is that the latter induces a *set* of partial orderings, corresponding to different assignments on its selector variables. Consider a conditionally acyclic TCP-net $\mathcal{N}$. The set of partial orders induced by $\mathcal{N}$ over its variables is consistent with the dependency graph of $\mathcal{N}$. In addition, if $\mathcal{S}(\mathcal{N})$ is the union of the selector variables in $\mathcal{N}$, then let $\mathcal{S}'(\mathcal{N}) \subseteq \mathcal{S}(\mathcal{N})$ be a *prefix* of $\mathcal{S}(\mathcal{N})$ if and only if, for each $X \in \mathcal{S}'(\mathcal{N})$, and for each $Y \in \mathcal{S}(\mathcal{N}) \setminus \mathcal{S}'(\mathcal{N})$, $X$ is not reachable from $Y$ in the dependency graph of $\mathcal{N}$. Observe, that any set of partial orders over the variables of $\mathcal{N}$, that agree on an assignment on a prefix $\mathcal{S}'(\mathcal{N})$ of $\mathcal{S}(\mathcal{N})$, agree on the ordering of all the variables in $\mathcal{N}$, the relative importance of which is fully determined by $\mathcal{S}'(\mathcal{N})$.

The Search algorithm is guided by the underlying TCP-net $\mathcal{N}$. It proceeds by assigning values to the variables in a top-down manner, assuring that outcomes are generated according to the preferential ordering induced by $\mathcal{N}$ – on a

---

Search $(\mathcal{N}, C, \mathcal{K})$
Input: Conditionally acyclic TCP-net $\mathcal{N}$, Constraints $C$,
　　　Context $\mathcal{K}$ (partial assignment on the original TCP-net)
Output: Set of all, Pareto-optimal w.r.t. $\mathcal{N}$, solutions for $C$.

Choose any variable $X$ s.t. there is no cp-arc $\langle \overrightarrow{Y, X} \rangle$,
　no i-arc $(\overline{Y, X})$, and no $(X, Y)$ in $\mathcal{N}$.
Let $x_1 \succ \ldots \succ x_k$ be the preference ordering of $\mathcal{D}(X)$
　given the assignment on $Pa(X)$ in $\mathcal{K}$.
Initialize the set of local results by $\mathcal{R} = \emptyset$
for $(i = 1;\ i \leq k;\ i + +)$ do
　$X = x_i$
　Strengthen the constraints $C$ by $X = x_i$ to obtain $C_i$
　if $C_j \subseteq C_i$ for some $j < i$ or $C_i$ is inconsistent then
　　continue with the next iteration
　else
　　Let $\mathcal{K}'$ be the partial assignment induced by $X = x_i$ and $C_i$
　　$\mathcal{N}_i$ = Reduce $(\mathcal{N}, \mathcal{K}')$
　　Let $\mathcal{N}_i^1, \ldots, \mathcal{N}_i^m$ be the components of $\mathcal{N}_i$, connected
　　　either by the edges of $\mathcal{N}_i$ or by the constraints $C_i$.
　　for $(j = 1;\ j \leq m;\ j + +)$ do
　　　$\mathcal{R}_i^j$ = Search $(\mathcal{N}_i^j, \mathcal{K} \cup \mathcal{K}', C_i)$
　　if $\mathcal{R}_i^j \neq \emptyset$ for all $j \leq m$ then
　　　foreach $o \in \mathcal{K}' \times \mathcal{R}_i^1 \times \cdots \times \mathcal{R}_i^m$ do
　　　　if for each $o' \in \mathcal{R}$ holds $\mathcal{K} \cdot o' \not\succ \mathcal{K} \cdot o$ then Add $o$ to $\mathcal{R}$
return $\mathcal{R}$

Reduce $(\mathcal{N}, \mathcal{K}')$
foreach $\{X = x_i\} \in \mathcal{K}'$ do
　foreach cp-arc $\langle \overrightarrow{X, Y} \rangle \in \mathcal{N}$ do
　　Restrict the CPT of $Y$ to the rows dictated by $X = x_i$.
　foreach ci-arc $\gamma = (Y_1, Y_2) \in \mathcal{N}$ s.t. $X \in \mathcal{S}(\gamma)$ do
　　Restrict the CIT of $\gamma$ to the rows dictated by $X = x_i$.
　　if, given the restricted CIT of $\gamma$, relative importance
　　　between $Y_1$ and $Y_2$ is independent of $\mathcal{S}(\gamma)$, then
　　　　if CIT of $\gamma$ is not empty then
　　　　　Replace $\gamma$ by the corresponding i-arc.
　　　　else Remove $\gamma$.
　Remove from $\mathcal{N}$ all the edges involving $X$. return $\mathcal{N}$.

Figure 2: The Search algorithm for TCP-nets.

---

call to the Search procedure with a TCP-net $\mathcal{N}$, the eliminated variable $X$ is one of the root variables of $\mathcal{N}$. The values of $X$ are considered according to the preferential ordering induced by the assignment on $Pa(X)$. Note that $X$ is observed in some context $\mathcal{K}$ which necessarily contains some assignment on $Pa(X)$. Whenever a variable $X$ is assigned to a value $x_i$, the current set of constraints $C$ is being strengthened into $C_i$. As a result of this propagation of $X = x_i$, values for some variables (at least for the variable $X$) will be fixed automatically, and this partial assignment $\mathcal{K}'$ will extend the current context $\mathcal{K}$ in processing of the next variable. The Reduce procedure refines the TCP-net $\mathcal{N}$ with respect to $\mathcal{K}'$: For each variable assigned by $\mathcal{K}'$, we reduce both the CPTs and the CITs involving this variable, and remove this variable from the network. This reduction of the CITs may remove conditioning of relative importance between some variables, and thus convert some ci-arcs into i-arcs, and/or to remove some ci-arcs completely. The central point is that, in contrast to CP-nets, for a pair of $X$-values $x_i, x_j$, the dependency graphs of the networks $\mathcal{N}_i$ and $\mathcal{N}_j$, accepted by propagating $C_i$ and $C_j$, respectively, may *disagree* on the ordering of some variables.

If the partial assignment $\mathcal{K}'$ causes the current CP-net to become disconnected with respect to both the edges of the network and the inter-variable constraints, then each con-nected component invokes an independent search. This is because optimization of the variables within such a component is independent of the variables outside that component. In addition, after strengthening the set of constraints $C$ by $X = x_i$ to $C_i$, some pruning is taking place in the search tree (see the **continue** instruction in the algorithm).[3] Therefore, the search is depth-first branch-and-bound, where the set of nondominated solutions generated so far is a proper subset of the required set of the Pareto-optimal solutions for the problem, and thus it corresponds to the current lower bound.

---

[2]An outcome $o$ is said to be Pareto-optimal with respect to some preference order $\succ$ and a set of outcomes $S$ if there is no other $o'$ such that $o' \succ o$.

When the potentially nondominated solutions for a particular subgraph are returned with some assignment $X = x_i$, each such solution is compared to all nondominated solutions involving more preferred (in the current context $\mathcal{K}$) assignments $X = x_j, j < i$. A solution with $X = x_i$ is added to the set of the nondominated solutions for the current subgraph and context if and only if it passes this non-domination test. Note that, from the semantics of the TCP-net, given the same context $\mathcal{K}$, a solution involving $X = x_i$ can not be preferred to a solution involving $X = x_j, j < i$. Thus, the generated global set $\mathcal{R}$ never shrinks.

If we are interested in getting *one* Pareto-optimal solution for the given set of constraints (which is usually the case), then we can output the *first* feasible outcome that is generated by Search. No dominance queries between pairs of outcomes are required because there is nothing to compare the first accepted solution with. If we are interested in getting *all*, or even *a few* Pareto-optimal solutions, then the efficiency of the dominance queries becomes an important factor in the entire complexity of the Search algorithm.

The dominance inference problem with respect to the TCP-nets can be also treated as a search for an improving flipping sequence, where the notion of flipping sequence is extended from this for the CP-nets.

**Definition 4** A sequence of outcomes

$$b = c_0 \prec c_1 \prec \cdots \prec c_{m-1} \prec c_m = a$$

is an *improving flipping sequence with respect to a TCP-net* $\mathcal{N}$ is and only if, for $0 \leq i < m$, either

1. *(CP-flips)* outcome $c_i$ is different from the outcome $c_{i+1}$ in the value of exactly one variable $X_j$, and $c_i[j] \prec c_{i+1}[j]$ given the (same) values of $Pa(X_j)$ in $c_i$ and $c_{i+1}$, or
2. *(I-flips)* outcome $c_i$ is different from the outcome $c_{i+1}$ in the value of exactly *two* variables $X_j$ and $X_k$, $c_i[j] \prec c_{i+1}[j]$ and $c_i[k] \succ c_{i+1}[k]$ given the (same) values of $Pa(X_j)$ and $Pa(X_k)$ in $c_i$ and $c_{i+1}$, and $X_j \triangleright X_k$ given $\mathcal{RI}(X_j, X_k, \mathbf{Z})$ and the (same) values of $\mathbf{Z}$ in $c_i$ and $c_{i+1}$.

---

[3] This pruning was introduced in [1]. See [1] for its explanation and justification.

Clearly, each value flip in such a flipping sequence is sanctioned by the TCP-net $\mathcal{N}$, and the CP-flips are exactly the flips allowed in CP-nets.

**Lemma 5** *Given a TCP-net $\mathcal{N}$, and two outcomes $a$ and $b$, $a \succ b$ is a consequence of $\mathcal{N}$ if and only if there is an improving flipping sequence from $b$ to $a$ with respect to $\mathcal{N}$.*

Various methods can be used to search for a flipping sequence, and at least some of the techniques, developed for this task with respect to CP-nets in [2, 4], can be extended for the TCP-net model.

## 6 CONCLUSIONS

We introduced the notions of absolute and conditional *relative importance* between pairs of variables and extended the CP-net model [2] to capture these preference statements. The extended model is called TCP-net. We identified a wide class of TCP-nets that are satisfiable – the class of conditionally acyclic TCP-nets. Finally, we showed how this subclass of TCP-nets can be used in preference-based constrained optimization. We refer the reader to the full version of this paper, where the relevance of the TCP-net model to the area of product configuration is discussed.

An important open theoretical question is the precise complexity of dominance testing in TCP-nets. Recent results in the context of CP-nets [3] do not seem immediately adaptable to TCP-nets. Finally, the question of consistency of TCP-nets that are not conditionally acyclic is another important challenge.